\documentclass[11pt]{article}

\usepackage[preprint]{acl}

\usepackage{times}
\usepackage{latexsym}

\usepackage[T1]{fontenc}

\usepackage[utf8]{inputenc}

\usepackage{microtype}

\usepackage{inconsolata}

\usepackage{graphicx}
\usepackage{hyperref}
\usepackage{url}
\usepackage{booktabs}
\usepackage{multirow}
\usepackage{graphicx}
\usepackage{array}
\usepackage{colortbl}
\usepackage{xcolor}
\usepackage{wrapfig}
\usepackage{graphicx}
\usepackage{subcaption}
\usepackage{natbib}
\usepackage{mathtools}
\usepackage{amsmath,amssymb}

\title{PDTrim: Targeted Pruning for Prefill-Decode Disaggregation in Inference}

\author{\bf Hao Zhang\thanks{These authors contribute equally to this work.},     Mengsi Lyu\footnotemark[1],
    Zhuo Chen,
    Xingrun Xing,
    Yulong Ao\thanks{Corresponding author.},
    Yonghua Lin\footnotemark[2] \\
    Beijing Academy of Artificial Intelligence (BAAI)}

\begin{document}
\maketitle

\begin{abstract}
Large Language Models (LLMs) demonstrate exceptional capabilities across various tasks, but their deployment is constrained by high computational and memory costs. Model pruning provides an effective means to alleviate these demands. However, existing methods often ignore the characteristics of prefill-decode (PD) disaggregation in practice. In this paper, we propose a pruning method that is highly integrated with PD disaggregation, enabling more precise pruning of blocks. Our approach constructs pruning and distillation sets to perform iterative block removal, obtaining better pruning solutions. Moreover, we analyze the pruning sensitivity of the prefill and decode stages and identify removable blocks specific to each stage, making it well suited for PD disaggregation deployment. Extensive experiments demonstrate our approach consistently achieves strong performance in both PD disaggregation and PD unified (non-PD disaggregation) settings, and can also be extended to other non-block pruning methods. Under the same settings, our method achieves improved performance and faster inference.
\end{abstract}

\section{Introduction}

Large Language Models (LLMs) have emerged as revolutionary tools, achieving state-of-the-art (SOTA) results across diverse tasks and applications \citep{ding2022, qin2023, zhu2023, li2023}. However, the rapid growth in model scale has posed significant challenges for their practical deployment \citep{zhang2023, choi2025retrospective, long2025sliminfer}. To address it, various techniques have been proposed, including pruning \citep{ma2023llm,ashkboos2024slicegpt,li2023communication,sun2025speed}, quantization \citep{liu2021post,zhou2021smoothquant,cai2023gptq,zhou2024framequant} and knowledge distillation \citep{hinton2015distilling,gou2021knowledge,yang2021knowledge,2024CDL}. Among them, pruning stands out as an effective strategy for reducing model size by eliminating redundant or less critical components, thereby lowering computational and storage costs.

While pruning offers clear benefits, its application to LLMs remains challenging. During inference, the prefill and decode stages are usually disaggregated (PD disaggregation) \citep{zhong2024distserve,patel2024splitwise,qin2024mooncake,dong2025hydrainfer}, allowing each stage to be optimized according to its specific resource requirements. Without such disaggregation, heterogeneous workloads must conform to a single resource profile, which results in suboptimal utilization and ultimately degrades performance. However, existing pruning methods often ignore the characteristics of PD disaggregation deployment in real deployments. In such systems, the prefill and decode stages exhibit markedly different sensitivities to pruning, making uniform strategies ineffective.

\begin{figure*}[t]
    \centering
    \includegraphics[width=1\linewidth]{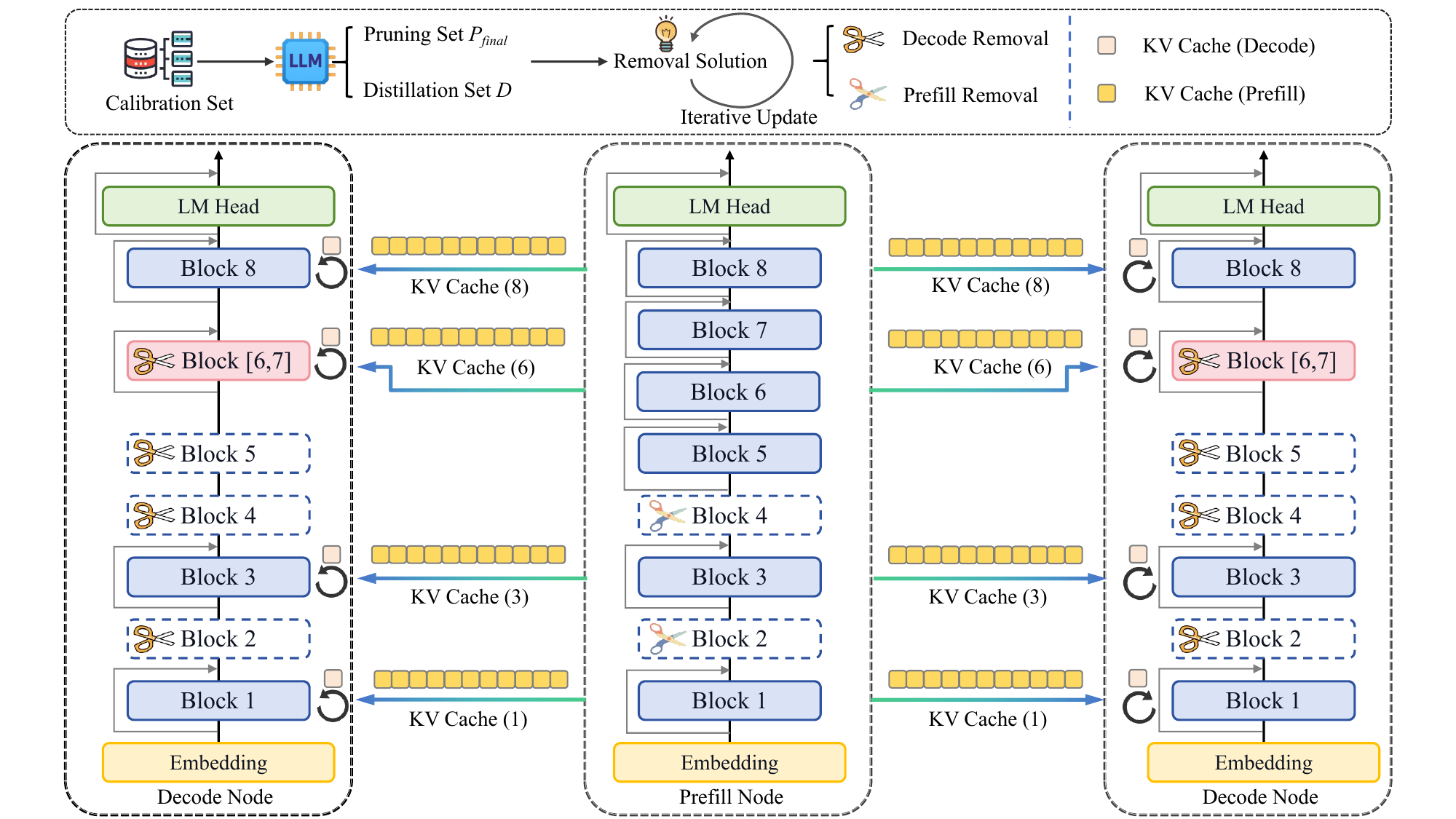}
    \caption{Overview of our pruning method combined with PD disaggregation. Under PD disaggregation, inputs are first processed on the prefill node and then dispatched to multiple decode nodes. The prefill node transfers the KV cache to the decode nodes, while each decode node also generates its own KV cache. In our method, we construct pruning and distillation sets and iteratively optimize the block removal based on these sets. The prefill and decode nodes apply different removal schemes. In this case, the prefill node prunes Blocks 2 and 4, whereas the decode nodes prune Blocks 2, 4 and 5, and further distill Blocks 6 and 7 into a single Block.}
    \label{fig:fig1}
\end{figure*}

In this paper, we propose a pruning method that is highly integrated with PD disaggregation, enabling more precise pruning of blocks. (1) Specifically, we introduce a approach for constructing pruning and distillation sets tailored to block pruning. (2) Building on these sets, we perform an iterative block removal procedure for LLMs. (3) We further analyze the pruning sensitivity of the prefill and decode stages separately, and identify removable blocks specific to each stage. This stage-aware pruning strategy aligns with the distinct execution characteristics of prefill and decode, resulting in particularly efficient PD disaggregation deployment. Compared to prior block pruning approaches, our method achieves better solutions. Extensive experiments demonstrate that our approach consistently achieves strong performance in both PD disaggregation and PD unified (non-PD disaggregation) settings. Our method can also be extended to other non-block pruning scenarios. Under the same settings, our method achieves improved performance and faster inference. In summary, our contributions can be summarized as follows:
\begin{itemize}
  \item We introduce a construction of pruning and distillation sets for block pruning, and leverage them to perform an iterative block removal strategy for LLMs. This can achieve superior performance over prior methods.
  \item We analyze the pruning sensitivity of the prefill and decode stages and identify removable blocks specific to each stage, making it well suited for PD disaggregation deployment.
  \item Our approach consistently achieves strong performance in both PD disaggregation and PD unified (non-PD disaggregation) settings, and can also be extended to other non-block pruning methods.
\end{itemize}

\section{Related Work}
\subsection{PD Disaggregation}
LLM inference can be divided into the prefill and decode stages. The prefill stage processes the entire input sequence to compute contextual representations and generate KV Cache. The decode stage then autoregressively generates tokens conditioned on previously generated outputs. PD disaggregation is a technique that explicitly decouples these two stages during inference, enabling deployment strategies tailored to the unique resource requirements of each stage. DistServe \citep{zhong2024distserve} is one of the seminal works in PD disaggregation. It assigns the prefill and decode stages to different GPUs, allowing for optimizations tailored to the unique characteristics of each stage. Splitwise \citep{patel2024splitwise} disaggregates the prompt computation and token generation stages and runs them on different machines. The TetriInfer \citep{hu2024inference} also leverages PD disaggregation to reduce interference between different downstream tasks. MemServe \citep{hu2024memserve} further explores memory optimization techniques within the PD disaggregation architecture. TaiChi \citep{wang2025prefill} integrates PD disaggregation and aggregation by leveraging differentiated GPU roles and adaptive scheduling to optimize goodput across diverse SLO regimes. Adrenaline \citep{liang2025injecting} offloads part of decode phase attention computation to prefill instances, improving LLM serving.


\subsection{Block Pruning}
To develop a straightforward pruning algorithm that is easy to deploy for LLMs, several studies have proposed removing less important blocks, a strategy that can lead to significant inference acceleration and improved computational efficiency. For example, methods \citep{men2024shortgpt} using cosine similarity to evaluate block importance combined with greedy pruning have been introduced. However, greedy approaches often fail to identify the globally optimal pruning configuration and tend to exhibit instability in practice. Techniques like LaCo \citep{yang2024laco} merge subsequent layers into preceding ones but typically sacrifice accuracy compared to direct layer removal. LLM-Streamline \citep{chen2024streamlining} reduces model size by distilling multiple consecutive blocks into a single block. However, overly aggressive distillation can significantly degrade model performance, and the approach still relies on a greedy block selection strategy. Other methods, such as SLEB \citep{song2024sleb} and Shortened LLaMA \citep{kim2024shortened}, adopt iterative pruning strategies guided by carefully designed importance metrics. However, these methods require recalculating importance scores using a calibration set after each block removal, resulting in substantial computational overhead. Consequently, developing a block pruning algorithm that is highly accurate remains a difficult and unresolved challenge.

\section{Methodology}




In this section, we describe our iterative block pruning method that is deeply integrated with PD disaggregation. Our goal is to select the \( k \) blocks whose removal minimally impacts performance. Figure \ref{fig:fig1} illustrates an overview of our method.

\subsection{Strategic Design of Sets for Pruning and Distillation}
We use a combined pruning and distillation strategy for refined block removal. Specifically, we first assess the redundancy of each block by calculating the cosine similarity \citep{men2024shortgpt} between its input and output (note that the similarity metric is not a core contribution of ours). We provide further details on this in Appendix \ref{Analysis of Using Cosine Similarity}. The redundancy of \(block_i\) is mathematically defined as follows:
\begin{equation}
\label{equation_r}
r_i = \cos(h_{i-1}, h_i)
\end{equation}
Here, \( r_i \) denotes the redundancy of \(block_i\) and the \(\cos\) measures the similarity between the input \( h_{i-1} \) and the output \( h_i \) of \(block_i\). Based on this redundancy metric, we directly identify the \( Top\left\lceil \frac{k}{2} \right\rceil \) blocks with the highest cosine similarity for removal. Given their high redundancy, these blocks are deemed suitable for direct pruning, thereby forming an initial pruning set of \( \left\lceil \frac{k}{2} \right\rceil \) blocks. The initial pruning set $\mathcal{P}_{initial}$ can be represented as follows:
\begin{equation}
\scalebox{0.83}{$
\mathcal{P}_{initial} = \{ block_i \mid r_i \in Top\left\lceil \tfrac{k}{2} \right\rceil \left( \{ r_1, r_2, \ldots, r_L \} \right) \}
$}
\end{equation}
For the blocks that are not included in the initial \( \left\lceil \frac{k}{2} \right\rceil \) blocks designated for pruning, we consider incorporating them into the construction of the distillation set. To ensure the effectiveness of the distillation, we limit distillation to pairs of consecutive blocks. Specifically, for a pair of consecutive blocks, denoted as \(block_i\) and \(block_{i+1}\), we define a metric \( d_i \) to determine whether they should be included in the distillation set. The metric \( d_i \) is defined as follows:
\begin{equation}
\begin{aligned}
d_i = \tfrac{1}{2} \left( \cos(h_{i-1}, h_{i+1}) + \max(
\cos(h_{i-1}, h_{i}),\cos(h_{i}, h_{i+1})) \right)
\end{aligned}
\end{equation}
This metric is designed from two complementary perspectives. Here, the \(\cos(h_{i-1}, h_{i+1})\) measures redundancy when treating two blocks as a single unit; higher values imply weaker transformation and easier merging. During distillation, we merge consecutive blocks by finetuning the block with lower redundancy and transferring representational capacity from the block with higher redundancy. The more redundant block is identified by \( \max( \cos(h_{i-1}, h_{i}), \cos(h_{i}, h_{i+1}))\); a larger value indicates higher redundancy and easier capacity transfer. We incorporate a pair of consecutive blocks into the distillation set and distill them into a single block only when the metric \( d_i \) exceeds a threshold, denoted as \( d_T \). The distillation set $\mathcal{D}$ can be represented as follows:
\begin{equation}
\begin{aligned}
    \mathcal{D} = \{ (block_i, block_{i+1}) \ | \ d_i \geq d_T \ {and} \\
    \ \{block_i, block_{i+1}\} \cap \mathcal{P}_{initial}  = \emptyset \}
\end{aligned}
\end{equation}
If a block meets the threshold with both its preceding and succeeding blocks, we select the one with the higher metric to include in the distillation set. Through this approach, we construct the distillation set, where each element represents the distillation of a pair of consecutive blocks into one block. Once the distillation set is constructed, all blocks not included in it are assigned to the pruning set. The final pruning set \(\mathcal{P}_{final}\) can be represented as follows:
\begin{equation}
\begin{aligned}
    \mathcal{P}_{final} = \{ block_i | \ \{(block_{i-1},block_{i}),\\ 
    (block_{i},block_{i+1})\} \cap \mathcal{D}  = \emptyset \}
\end{aligned}
\end{equation}
In this setup, each block in the model is assigned to either the pruning set or the distillation set. Ingeniously, the number of blocks removed by each element in the distillation set matches that in the pruning set, with both strategies removing exactly one block. This consistency establishes a foundation for iterative optimization strategies.

\subsection{Iterative Optimization toward Optimal Block Removal}
To determine the optimal combination of block removals, we iteratively refine the removal set to minimize the impact on model performance. Specifically, we first initialize the block removal set by selecting the \( Top\left\lceil \frac{k}{2} \right\rceil \) blocks with the highest cosine similarity from the pruning set. The remaining \( k -\left\lceil \frac{k}{2} \right\rceil \) blocks \(\mathcal{B}_{partial}\) are chosen from the distillation set based on the aforementioned metrics \(d_i\) in descending order. If the number of blocks in the distillation set is insufficient, we supplement them from the pruning set in descending order of cosine similarity, excluding the \( Top\left\lceil \frac{k}{2} \right\rceil \) blocks that need to be pruned. The initial block removal set \(\mathcal{B}_{initial}\) can be represented as follows:

\begin{equation}
\mathcal{B}_{partial}=
\left\{
\begin{aligned}
& {Top}(k - \left\lceil \frac{k}{2} \right\rceil)(\mathcal{D}) \\
& \mbox{\scriptsize$\bigl(
{if } \ |\mathcal{D}| \geq k - \left\lceil \frac{k}{2} \right\rceil
\bigr)$} \\[4pt]
& \scalebox{0.8}{$
\mathcal{D} \cup {Top}\left(k - \left\lceil \frac{k}{2} \right\rceil - |\mathcal{D}|\right)
\left(\mathcal{P}_{final} \setminus \mathcal{P}_{initial}\right)
$} \\
& \mbox{\scriptsize$\bigl(
{if } \ |\mathcal{D}| < k - \left\lceil \frac{k}{2} \right\rceil
\bigr)$}
\end{aligned}
\right.
\end{equation}

\begin{equation}
    \mathcal{B}_{initial} = \mathcal{P}_{initial} \ \cup \ \mathcal{B}_{partial}
\end{equation}
We define several key parameters for the optimization process. The initial temperature \( T_0 \) controls the randomness in the block iteration, the decay coefficient \( \alpha \) is employed to gradually reduce the temperature and the termination temperature \( T_{min} \) determines when to stop the entire block iteration (we provide the analysis of these parameters in experiments). In each iteration, we select an element from the unremoved set based on probability to replace a random element in the current removed set, thereby generating a new neighborhood solution.
To determine the probability of each element being selected, we assign a weight \( \omega_i \) to each element \( i \) in the unremoved set. For every element \( i \), \( block_i \) in the pruning set or element \( block_i \) and \( block_{i+1} \) in the distillation set, the weight \( \omega_i \) and the probability \( p_i \) are related to formula \ref{equation_r}, being selected can be defined as follows: 
\begin{equation}
\begin{aligned}
\omega_i &=
\begin{cases}
r_i & \textit{pruning set} \\
\frac{1}{2}(r_i + r_{i+1}) & \textit{distillation set}
\end{cases}
\end{aligned}
\end{equation}
\begin{equation}
    p_i = \frac{\omega_i}{\sum_{j=1}^{N} \omega_j}
\end{equation}
{Here, \( N \) denotes the total number of elements in the unremoved set. For the new candidate block removal solution \(s'\), we calculate its accuracy on the calibration set as the objective function value \( f_{s'} \), and compare it with the current solution's (\(s\)) objective function value \( f_{s} \). If the accuracy of the candidate solution is higher ( \(f_{s'}>f_{s}\) ), we directly accept it. However, if the candidate solution's accuracy is lower ( \(f_{s'}<f_{s}\) ), we decide whether to accept it based on a probability. The acceptance probability \(P_{{s'}}\) is calculated using the following formula:}
\begin{equation}
P_{{s'}} = e^{-\frac{\Delta f}{T}} \quad \Delta f = f_{{s}} - f_{{s'}}
\end{equation}
where \( \Delta f\) is the difference in objective function values between the current solution and the new candidate solution, and \( T \) is the current temperature. This probability based acceptance mechanism allows us to escape local optima and explore the solution space more comprehensively. After evaluating the new solution, we update the temperature using the decay factor \(T = \alpha T\). We repeat this process until the temperature reaches the minimum threshold \( T_{{min}} \). Our experiments show that even with fewer iterations, performance still remains strong. Throughout the entire iterative process, we continuously record the block removal combination with the highest performance, considering it as the optimal solution. The optimal solution \(\mathcal{B}_{{optimal}}\) can be represented as follows:
\begin{equation}
    \mathcal{B}_{{optimal}} = \arg\max_{\mathcal{B} \in {solutions}} {f}_{\mathcal{B}}
\end{equation}
where \( {f}_{\mathcal{B}} \) is the objective function value of solution \({\mathcal{B}}\). This strategy yields better solutions that approximate the global optimum, as confirmed by experiments. It requires only a few iterations to achieve improved results, and the partial randomness introduced does not noticeably degrade performance. Moreover, it incurs substantially lower pruning overhead compared to other global search methods, as verified empirically.

\subsection{Heterogeneous Strategies Across Prefill and Decode Nodes}
\label{Different Strategies for Prefill and Decode}
The prefill stage is more sensitive to pruning than the decode, pruning during this stage tends to lead to a more significant drop in performance. Here, we provide an abstract analysis, ignoring other complex effects. Denote by $X \in \mathbb{R}^{N \times d}$ the input sequence in the prefill stage and by $x_t \in \mathbb{R}^{1 \times d}$ the initial input in the decode stage. We denote the perturbations of the attention parameters ${W}_Q$, ${W}_K$ and ${W}_V$ by $\Delta {W}_Q$, $\Delta {W}_K$ and $\Delta {W}_V$, respectively. The query vector \(q\) and its perturbation \(\Delta q\) for \(x_t\) are given by \(q = x_t W_Q\) and \(\Delta q = x_t \Delta W_Q\). The concatenated key and value matrices are defined as \((K,V)\), where \(K = [K_{pre},\, x_t W_K]\) and \(V = [V_{pre},\, x_t W_V]\), with corresponding perturbations \(\Delta K = [\Delta K_{pre},\, x_t \Delta W_K]\) and \(\Delta V = [\Delta V_{pre},\, x_t \Delta W_V]\).
Here, \( K_{{pre}} = X W_K \) and \( \Delta K = X \Delta W_K \); the same applies to the others. Then, the attention output before \((o)\) and after perturbation \((\tilde{o})\), as well as the resulting error \((E)\), can be expressed as follows:
\begin{equation}
o = {softmax}\left( \tfrac{q K^\top}{\sqrt{d}} \right) V
\end{equation}
\begin{equation}
\scalebox{0.95}{$
    \tilde{o} = {softmax}\\
\left( \tfrac{(q + \Delta q) (K + \Delta K)^\top}{\sqrt{d}} \right) (V + \Delta V)
$}
\end{equation}
\begin{equation}
\tilde{A} = {softmax}\left(\tfrac{(q + \Delta q)(K + \Delta K)^\top}{\sqrt{d}}\right)
\end{equation}
\begin{equation}
    \Delta A = \tilde{A} - {softmax}\left(\tfrac{q K^\top}{\sqrt{d}}\right)
\end{equation}
\begin{equation}
E = \tilde{o} - o 
= {\Delta A \, V} + {\tilde{A} \, \Delta V}
\end{equation}
Considering the above analysis, we obtain the following formulas \textit{(The detailed derivation of these formulas can be found in Appendix \ref{math_setp} and \ref{softmax}):}
\begin{equation}
\begin{aligned}
\|E\|_F
&\le \frac{L_{{softmax}}}{\sqrt{d}}
\Big(
\|\Delta q\|_2 \, \|K\|_F
+ \|q\|_2 \, \\
&\|\Delta K\|_F
\Big)\cdot \|V\|_F  + \|\Delta V\|_F 
\end{aligned}
\end{equation}
\begin{equation}
E = {G}(x_t, Y_{{pre}}, \Delta)
\end{equation}
Here, $L_{{softmax}}$ denotes the Lipschitz constant of the softmax, $\|E\|_F$ denotes the Frobenius norm of the error $E$, and $G$ is an abstract error function. $Y_{{pre}}$ denotes the set of all results (including both intermediate and final outputs) produced in the previous step that can be reused in the current step, and $\Delta$ represents the set of perturbations. From this, we observe that pruning errors accumulate as the sequence is generated. During the prefill stage, a perturbation simultaneously contaminates the representations of all $N$ tokens. These corrupted representations are repeatedly accessed in subsequent decode steps, causing the error to accumulate and amplify over time. In contrast, a single decode step only affects the current input and its subsequent states, limiting the scope of impact. {Therefore, after determining the optimal combination of removed blocks, we further adjust the removed blocks for each stage. Specifically, for each element in the current combination of removed blocks, we test it on the calibration set. If removing only in the decode stage can significantly improve performance, then that element is removed only in the decode stage while being retained in the prefill stage. If there is no significant performance improvement, then the removing operation for that element will be applied in both the prefill and decode stages. More details can be found in Appendix \ref{More_Details_on_Prefill_and_Decode_Removal_Schemes} and \ref{Experimental Settings}. This approach allows us to determine the removed blocks for the prefill and the decode separately.}

\begin{table*}[]
\centering
\setlength{\tabcolsep}{4pt} 
\resizebox{\textwidth}{!}{%
\begin{tabular}{c|c|ccccccccccccc|c}
\toprule
\textbf{LLM} & \textbf{Method} & \textbf{MMLU} & \textbf{CMMLU} & \textbf{ARC-E} & \textbf{ARC-C} & \textbf{PIQA} & \textbf{Winog} & \textbf{HSwag} & \textbf{BoolQ} & \textbf{MathQA} & \textbf{WNLI} & \textbf{SST-2} & \textbf{RTE} & \textbf{CB} & \textbf{AVG} \\ 
\midrule
\multirow{7}{*}{\rotatebox[origin=c]{90}{\textbf{LLaMA3.1-8B}}} 
& Dense      & 63.35 & 50.85 & 81.52 & 51.28 & 80.30 & 74.03 & 60.05 & 82.26 & 39.56 & 59.15 & 76.83 & 71.12 & 60.71 & 65.46 \\
& LLM-Pruner & 52.01 & 41.12 & 67.36 & 42.06 & 74.54 & 69.32 & 51.50 & 73.78 & 32.35 & 50.78 & 69.37 & 49.40 & 55.28 & 56.07 \\
& FLAP       & 52.11 & 41.13 & 67.42 & 42.15 & 74.63 & 69.36 & 51.56 & \textbf{73.82} & 32.39 & 50.82 & 69.41 & 49.44 & 55.34 & 56.11 \\
& Shortened  & 33.54 & 34.28 & 72.31 & 42.15 & 72.63 & 69.61 & 47.96 & 45.23 & 34.27 & 56.34 & 52.52 & 67.15 & 69.64 & 53.66 \\
& ShortGPT   & 51.92 & 41.11 & 67.34 & 42.06 & 74.48 & 69.38 & 51.56 & 73.76 & 32.50 & 50.70 & 69.31 & 49.50 & 55.36 & 56.08 \\
& SLEB       & 28.35 & 25.51 & 71.04 & 36.18 & 75.46 & 62.98 & 49.70 & 57.89 & 27.30 & 46.48 & 55.85 & 57.04 & 35.71 & 48.42 \\
& Ours & \textbf{60.82} & \textbf{46.66} & \textbf{73.15} & \textbf{43.86} & \textbf{76.99} & \textbf{73.56} & \textbf{53.80} & 69.85 & \textbf{34.28} & \textbf{60.56} & \textbf{83.03} & \textbf{70.76} & \textbf{73.21} & \textbf{62.99} \\
\midrule
\multirow{7}{*}{\rotatebox[origin=c]{90}{\textbf{LLaMA2-13B}}} 
& Dense      & 52.10 & 34.73 & 79.42 & 48.38 & 79.05 & 72.22 & 60.07 & 80.61 & 32.09 & 66.20 & 87.61 & 69.31 & 80.36 & 64.78 \\
& LLM-Pruner & 50.11 & 33.57 & 61.16 & 37.67 & 71.38 & 70.71 & 47.37 & 62.53 & 24.41 & 43.24 & 65.37 & 59.19 & \textbf{51.38} & 52.16 \\
& FLAP       & 49.89 & 33.91 & 60.90 & 37.65 & 71.44 & 70.69 & 47.54 & 62.43 & 24.78 & 43.26 & 64.91 & 59.19 & 51.34 & 52.15 \\
& Shortened  & 26.71 & 25.50 & 26.26 & 22.61 & 51.14 & 48.22 & 25.76 & 38.93 & 18.89 & 43.66 & 46.56 & 52.71 & 37.50 & 35.73 \\
& ShortGPT   & 50.13 & 33.97 & 61.32 & 37.88 & 71.44 & \textbf{70.80} & 47.71 & 62.54 & 24.82 & 43.66 & 65.37 & 59.57 & 51.79 & 52.38 \\
& SLEB       & 23.76 & 25.41 & 67.85 & 33.87 & \textbf{75.41} & 63.77 & 48.76 & 62.42 & 25.46 & 45.07 & 50.92 & 58.84 & 41.07 & 47.89 \\
& Ours & \textbf{51.52} & \textbf{35.37} & \textbf{69.78} & \textbf{40.02} & 74.43 & 70.72 & \textbf{53.11} & \textbf{62.84} & \textbf{28.44} & \textbf{53.52} & \textbf{80.16} & \textbf{64.26} & {46.43} & \textbf{56.20} \\
\midrule
\multirow{7}{*}{\rotatebox[origin=c]{90}{\textbf{Qwen2.5-7B}}} 
& Dense      & 71.92 & 81.69 & 80.56 & 48.21 & 78.73 & 72.93 & 59.92 & 84.46 & 43.22 & 71.83 & 91.86 & 81.23 & 87.50 & 73.39 \\
& LLM-Pruner & 36.96 & 36.71 & 71.33 & 38.20 & 74.86 & 55.79 & 47.71 & 55.67 & 31.08 & 60.43 & 70.79 & 54.42 & 38.81 & 51.75 \\
& FLAP       & 37.17 & 36.39 & 71.47 & 37.83 & 76.21 & 55.65 & 47.47 & 57.23 & 30.11 & 61.61 & 70.68 & 54.59 & 37.32 & 51.82 \\
& Shortened  & 24.90 & 25.08 & 25.25 & 20.31 & 53.65 & 50.99 & 25.69 & 37.83 & 19.50 & 53.52 & 50.92 & 46.21 & 19.64 & 34.88 \\
& ShortGPT   & 36.89 & 31.07 & 71.46 & 37.80 & 76.12 & 55.80 & 48.67 & 63.21 & 30.59 & 42.25 & \textbf{80.62} & 54.87 & 41.07 & 51.57 \\
& SLEB       & 38.56 & 38.25 & \textbf{71.84} & \textbf{39.42} & 76.77 & 55.96 & 47.98 & 57.49 & \textbf{31.12} & 61.97 & 72.25 & 56.32 & 39.29 & 52.86 \\
& Ours & \textbf{52.96} & \textbf{51.34} & 71.09 & 39.16 & \textbf{77.04} & \textbf{56.99} & \textbf{51.19} & \textbf{72.20} & 30.95 & \textbf{66.20} & 63.53 & \textbf{65.34} & \textbf{57.14} & \textbf{58.09} \\
\midrule
\multirow{7}{*}{\rotatebox[origin=c]{90}{\textbf{Qwen2.5-14B}}} 
& Dense      & 77.45 & 84.44 & 82.37 & 56.31 & 81.12 & 75.37 & 63.37 & 85.23 & 53.03 & 77.46 & 89.11 & 79.78 & 80.36 & 75.80 \\
& LLM-Pruner & 43.09 & 42.01 & 73.33 & 40.54 & 73.58 & 58.47 & 47.79 & 61.89 & 31.53 & 48.79 & 54.26 & 56.88 & 49.27 & 52.42 \\
& FLAP       & 44.86 & 44.77 & 50.29 & 31.00 & 60.72 & 51.44 & 34.14 & 65.16 & 25.65 & 65.28 & 89.95 & 76.87 & 68.74 & 54.53 \\
& Shortened  & 24.63 & 25.31 & 25.04 & 20.14 & 52.88 & 50.43 & 25.69 & 37.92 & 18.93 & 52.11 & 48.62 & 51.99 & 37.50 & 36.25 \\
& ShortGPT   & 45.75 & 45.63 & 50.63 & 31.40 & 61.81 & 52.64 & 34.41 & \textbf{65.72} & 26.16 & \textbf{66.20} & \textbf{91.28} & \textbf{78.34} & 69.64 & 55.35 \\
& SLEB       & 43.77 & 42.90 & \textbf{74.07} & 41.04 & \textbf{74.37} & 58.96 & \textbf{48.43} & 62.57 & 32.23 & 49.30 & 54.93 & 57.40 & 50.00 & 53.07 \\
& Ours & \textbf{72.01} & \textbf{76.82} & 68.81 & \textbf{44.37} & 71.38 & \textbf{70.17} & 47.98 & 64.80 & \textbf{34.91} & {61.97} & 82.11 & 74.01 & \textbf{69.65} & \textbf{64.54} \\
\bottomrule
\end{tabular}%
}
\caption{Performance comparison of pruning methods in LLMs across a variety of benchmarks. To increase the diversity of evaluation, we apply pruning rates of 13.6\%, 24.4\%, 15.3\% and 18.6\% for the four models, corresponding to commonly used pruning ratios for structured pruning.}
\label{tab:main_result}
\end{table*}

\section{Experiments}
\label{Experiments}
\subsection{Experimental Setup}
\label{main Experimental Settings}
\textbf{Models and Benchmarks }To demonstrate the effectiveness of our method, we conduct extensive evaluations on representative LLMs with diverse architectures and scales, including LLaMA3.1-8B \citep{grattafiori2024llama}, LLaMA2-13B \citep{touvron2023llama},  Qwen2.5-7B \citep{yang2024qwen2} and Qwen2.5-14B \citep{yang2024qwen2}. We also use Qwen2.5-32B \citep{yang2024qwen2} and OPT-6.7B \citep{zhang2022opt} in the supplementary experiments. We employ a wide range of benchmarks. These benchmarks include MMLU \citep{hendrycks2020mmlu}, CMMLU \citep{li2023cmmlu}, PIQA \citep{bisk2020piqa}, Winogrande \citep{ai2:winogrande}, HellaSwag \citep{zellers2019hellaswag}, BoolQ \citep{clark2019boolq}, MathQA \citep{amini2019mathqa}, ARC-Easy and ARC-Challenge \citep{clark2018ARC}, RTE \citep{wang2018glue}, WNLI \citep{wang2018glue}, CB \citep{wang2019superglue} and SST-2 \citep{wang2018glue}. This comprehensive protocol ensures thorough assessment.


\textbf{Baselines }We conduct comparative evaluations against other methods, including LLM-Pruner \citep{ma2023llm}, FLAP \citep{an2024fluctuation}, Shortened LLaMA  (abbreviated as Shortened)\citep{kim2024shortened}, ShortGPT \citep{men2024shortgpt} and SLEB \citep{song2024sleb}. Additionally, we also use SliceGPT \citep{ashkboos2024slicegpt} to test inference speed. We implement PD disaggregation versions of the classic channel pruning LLM-Pruner and block pruning ShortGPT according to our algorithm, and use them as additional baselines. More details can be found in Appendix \ref{Experimental Settings}.

\textbf{Implementation Details }Our experiments are conducted using the PyTorch framework \citep{paszke2019pytorch} and the Hugging Face Transformers library \citep{wolf2020transformers}. We use two nodes, each equipped with one NVIDIA H100 80GB GPU. When distilling two blocks, we use the weights of the block with low cosine similarity as the initial weights. By default, we use 256 randomly sampled examples from PIQA  and MMLU as the calibration set. The hyperparameter settings and training details can be found in Appendix \ref{Experimental Settings}.

\begin{table*}[t]
\centering
\resizebox{\textwidth}{!}{%
\begin{tabular}{c|c|cccccccc|c}
\toprule
\textbf{LLM} & \textbf{Method} & \textbf{MMLU} & \textbf{CMMLU} & \textbf{ARC-E} & \textbf{ARC-C} & \textbf{HSwag} & \textbf{MathQA} & \textbf{WNLI} & \textbf{RTE} & \textbf{AVG} \\
\midrule
\multirow{5}{*}{\textbf{LLaMA3.1-8B}} & LLM-Pruner & 52.01 & 41.12 & 67.36 & 42.06 & 51.50 & 32.35 & 50.78 & 49.40 & 48.32 \\
 & ShortGPT & 51.92 & 41.11 & 67.34 & 42.06 & 51.56 & 32.50 & 50.70 & 49.50 & 48.34 \\
\cmidrule(lr){2-11}
 & LLM-Pruner (Ours) & 54.95 & 44.30 & 72.46 & 43.82 & 52.84 & 34.13 & 53.94 & 52.88 & 51.16 \\
 & ShortGPT (Ours) & 54.62 & 43.89 & 71.38 & 43.11 & 53.55 & \textbf{35.08} & 53.62 & 52.63 & 50.98 \\
 & Ours & \textbf{60.82} & \textbf{46.66} & \textbf{73.15} & \textbf{43.86} & \textbf{53.80} & 34.28 & \textbf{60.56} & \textbf{70.76} & \textbf{55.49} \\
\midrule
\multirow{5}{*}{\textbf{LLaMA2-13B}} & LLM-Pruner & 50.11 & 33.57 & 61.16 & 37.67 & 47.37 & 24.41 & 43.24 & 59.19 & 44.59 \\
 & ShortGPT & 50.13 & 33.97 & 61.32 & 37.88 & 47.71 & 24.82 & 43.66 & 59.57 & 44.88 \\
\cmidrule(lr){2-11}
 & LLM-Pruner (Ours) & 52.58 & 34.27 & 65.55 & 39.19 & 49.99 & 25.96 & 44.53 & 62.54 & 46.83 \\
 & ShortGPT (Ours) & \textbf{53.70} & 34.28 & 64.25 & 38.98 & 50.76 & 26.50 & 46.84 & 63.41 & 47.34 \\
 & Ours & 51.52 & \textbf{35.37} & \textbf{69.78} & \textbf{40.02} & \textbf{53.11} & \textbf{28.44} & \textbf{53.52} & \textbf{64.26} & \textbf{49.50} \\
\midrule
\multirow{5}{*}{\textbf{Qwen2.5-7B}} & LLM-Pruner & 36.96 & 36.71 & 71.33 & 38.20 & 47.71 & 31.08 & 60.43 & 54.42 & 47.11 \\
 & ShortGPT & 36.89 & 31.07 & 71.46 & 37.80 & 48.67 & 30.59 & 42.25 & 54.87 & 44.20 \\
\cmidrule(lr){2-11}
 & LLM-Pruner (Ours) & 38.99 & 39.42 & 72.41 & 38.66 & 50.54 & \textbf{33.34} & 61.89 & 56.83 & 49.01 \\
 & ShortGPT (Ours) & 39.51 & 33.01 & \textbf{73.04} & 39.01 & 51.18 & 32.88 & 62.96 & 56.64 & 48.53 \\
 & Ours & \textbf{52.96} & \textbf{51.34} & 71.09 & \textbf{39.16} & \textbf{51.19} & 30.95 & \textbf{66.20} & \textbf{65.34} & \textbf{53.53} \\
\midrule
\multirow{5}{*}{\textbf{Qwen2.5-14B}} & LLM-Pruner & 43.09 & 42.01 & 73.33 & 40.54 & 47.79 & 31.53 & 48.79 & 56.88 & 48.00 \\
 & ShortGPT & 45.75 & 45.63 & 50.63 & 31.40 & 34.41 & 26.16 & 66.20 & 78.34 & 47.32 \\
\cmidrule(lr){2-11}
 & LLM-Pruner (Ours) & 65.75 & 45.32 & 74.49 & 43.13 & 48.75 & 32.43 & 51.27 & 58.44 & 52.45 \\
 & ShortGPT (Ours) & 68.34 & 47.36 & 52.21 & 33.16 & 36.37 & 28.05 & \textbf{67.64} & \textbf{78.89} & 51.50 \\
 & Ours & \textbf{72.01} & \textbf{76.82} & 68.81 & \textbf{44.37} & 47.98 & \textbf{34.91} & 61.97 & 74.01 & \textbf{60.11} \\
\bottomrule
\end{tabular}%
}
\caption{Performance comparison of our method with LLM-Pruner and ShortGPT, as well as their variants. LLM-Pruner (Ours) and ShortGPT (Ours) are our PD disaggregation extensions.}
\label{tab:llm-pruner_PD}
\end{table*}

\subsection{Main Results}
We evaluate our method against strong baselines on four representative LLMs and multiple benchmarks under identical experimental settings, including the calibration set. To diversify the evaluation, we adopt pruning ratios of 13.6\%, 24.4\%, 15.3\% and 18.6\% across the four models. The {Dense} configuration corresponds to the original uncompressed model, serving as a reference prior to pruning. The experiments are conducted with consistent parameter settings and implementation details. As shown in Table~\ref{tab:main_result}, our method consistently outperforms all baselines across different models. These results clearly confirm the effectiveness and robustness of our method under diverse pruning scenarios.

Our approach serves as an optimization for PD disaggregation and can be readily combined with other pruning methods. We extend two representative approaches, the channel pruning method LLM-Pruner and the block pruning method ShortGPT. Specifically, we first use LLM-Pruner and ShortGPT to select pruning weights for the decode stage, and then apply our method to identify a pruned subset of weights for the prefill stage. The resulting approaches are denoted as LLM-Pruner (Ours) and ShortGPT (Ours), respectively. As shown in Table~\ref{tab:llm-pruner_PD}, both methods exhibit substantial performance improvements on the evaluated datasets after applying PD disaggregation. Moreover, our approach consistently achieves the best overall results. These findings highlight the effectiveness of PD disaggregation as a general principle for enhancing pruning strategies in LLMs. To validate the scalability and generality of our method, we conduct experiments on larger model, additional datasets, and an extra metric (perplexity), achieving promising results (see  Appendix~\ref{Larger Model, More Datasets and Additional Metric}).

\subsection{Efficiency Analysis}





\paragraph{Inference Latency}We measure the execution time for a single inference of the LLaMA3.1-8B model with an input tensor of shape $[1,1024]$ in FP32 format. Specifically, we evaluate the effects of pruning 13.58\% and approximately 50\% of the parameters in LLaMA3.1-8B. We then compare our proposed method against several other widely used pruning algorithms, including SliceGPT, LLM-Pruner and FLAP. As shown in Table~\ref{tab:runtime}, our approach consistently achieves the fastest runtime, significantly reducing inference latency. This demonstrates that our block pruning approach enables faster inference speed.

\begin{table}[h!]
\centering
\resizebox{0.48\textwidth}{!}{ 
\begin{tabular}{c|ccccc}
\toprule
Method & Dense & SliceGPT & LLM-Pruner & FLAP & Ours \\
\midrule
Time (13.58\%) & 287.35 & 263.35 & 278.19 & 259.81 & \textbf{236.63} \\
\midrule
Time ($\sim$50\%) & 287.35 & 202.69 & 252.69 & 184.16 & \textbf{151.89} \\
\bottomrule
\end{tabular}
}
\caption{Comparison of inference time (ms) for the LLaMA3.1-8B on different pruning methods.}
\label{tab:runtime}
\end{table}

\paragraph{Pruning Overhead}We conduct a comparison of the computational overhead between our proposed pruning method and SLEB. SLEB \citep{song2024sleb} is also a pruning strategy that aims to identify the optimal pruning solution. We prune five blocks on LLaMA3.1-8B. The hyperparameters we use have been provided in the experiment details in the previous section. As shown in Table~\ref{tab:load}, our method achieves an average runtime of only 44.06 seconds, whereas SLEB requires 183.6 seconds on average. Moreover, we conduct a comparison of pruning overhead on LLaMA2-13B under a pruning rate of 24.4\%, where our approach continues to demonstrate a significant advantage. This efficiency improvement is attributed to our iterative pruning strategy, which can more rapidly converge to an effective pruning configuration.

\begin{table}[h!]
    \centering
    \small
    \resizebox{0.48\textwidth}{!}{
    \begin{tabular}{c|c|c}
\toprule
Model & Method &Pruning Time (s) \\
\midrule
\multirow{2}{*}{LLaMA3.1-8B} & SLEB \citep{song2024sleb} & 183.6 \\
 & Ours & 44.06 \\
\midrule
\multirow{2}{*}{LLaMA2-13B} & SLEB \citep{song2024sleb} & 858.12 \\
 & Ours & 96.52 \\
\bottomrule
\end{tabular}}
    \caption{Comparison of pruning time between our method and SLEB on LLaMA series models.}
    \label{tab:load}
\end{table}

\begin{table*}[htbp]
\centering
\resizebox{\textwidth}{!}{%
\begin{tabular}{c|cccccccc|c}
\toprule
\textbf{Method} & \textbf{MMLU} & \textbf{CMMLU} & \textbf{ARC-E} & \textbf{ARC-C} & \textbf{HSwag} & \textbf{MathQA} & \textbf{WNLI} & \textbf{RTE} & \textbf{AVG} \\
\midrule
Dense        & 52.10 & 34.73 & 79.42 & 48.38 & 60.07 & 32.09 & 66.20 & 69.31 & 55.29 \\
LLM-Pruner   & 50.11 & 33.57 & 61.16 & 37.67 & 47.37 & 24.41 & 43.24 & 59.19 & 44.59 \\
FLAP         & 49.89 & 33.91 & 60.90 & 37.65 & 47.54 & 24.78 & 43.26 & 59.19 & 44.64 \\
Shortened    & 26.71 & 25.50 & 26.26 & 22.61 & 25.76 & 18.89 & 43.66 & 52.71 & 30.26 \\
ShortGPT     & 50.13 & 33.97 & 61.32 & \textbf{37.88} & 47.71 & 24.82 & 43.66 & 59.57 & 44.88 \\
SLEB         & 23.76 & 25.41 & \textbf{67.85} & 33.87 & 48.76 & 25.46 & 45.07 & 58.84 & 41.13 \\
{Ours (Unified)} & \textbf{51.05} & \textbf{33.99} & {64.02} & \textbf{37.88} & \textbf{48.83} & \textbf{26.06} & \textbf{53.12} & \textbf{63.54} & \textbf{47.31} \\
\bottomrule
\end{tabular}%
}
\caption{Evaluation of pruning performance in the PD Unified scenario with LLaMA2-13B.}
\label{appendix_PDUnified}
\end{table*}
\begin{table*}[h!]
\centering
\resizebox{\textwidth}{!}{
\begin{tabular}{c|cccccccc|c}
\toprule
\textbf{Strategy} & \textbf{MMLU} & \textbf{CMMLU} & \textbf{ARC-E} & \textbf{ARC-C} & \textbf{HSwag} & \textbf{MathQA} & \textbf{WNLI} & \textbf{RTE} & \textbf{AVG} \\
\midrule
{Unified} & 51.05 & 33.99 & 64.02 & 37.88 & 48.83 & 26.06 & 53.12 & 63.54 & 47.31 \\
{Disaggregation} & \textbf{51.52} & \textbf{35.37} & \textbf{69.78} & \textbf{40.02} & \textbf{53.11} & \textbf{28.44} & \textbf{53.52} & \textbf{64.26} & \textbf{49.50} \\
\bottomrule
\end{tabular}
}
\caption{Performance comparison between PD Unified and PD Disaggregation on LLaMA2-13B.}
\label{PD}
\end{table*}

\subsection{Evaluation in the PD Unified Setting}
\label{Performance Comparison in the Non-PD Disaggregation Setting}

We compare the performance of our method with other approaches in the non-PD disaggregation setting, referred to as PD Unified. In this setting, the prefill and decode stages are not explicitly disaggregated during inference. Specifically, we evaluate our method on the LLaMA2-13B model with 24.4\% of its parameters pruned across multiple benchmarks. As shown in Table~\ref{appendix_PDUnified}, our approach continues to achieve strong and consistent performance in the PD Unified scenario. These results demonstrate that, even without explicit disaggregation of the prefill and decode stages, our method maintains performance across diverse tasks.

\subsection{Comparison between Disaggregation and Unified}
\label{Disaggregation_and_Unified_v1}
To validate the effectiveness of our strategy, we conduct extensive benchmarks on LLaMA2-13B by pruning 24.4\% of the model parameters. As a baseline, PD Unified applies the same parameter removal configuration to both the prefill and decode stages, without accounting for their differing sensitivities to pruning. As shown in Table~\ref{PD}, our PD disaggregation compression strategy consistently outperforms the PD unified baseline across benchmarks. These results provide strong evidence for the effectiveness of PD disaggregation pruning.

\subsection{Orthogonal to Quantization}
To evaluate the compatibility and effectiveness of our method under combined compression strategies, we further integrate it with quantization techniques. Specifically, we apply our approach to the Qwen2.5-32B with a pruning ratio of 25\% in conjunction with 8-bit AWQ quantization \citep{lin2024awq}. As shown in Table~\ref{quant}, our method consistently demonstrates robust performance, indicating that the pruning strategy is largely orthogonal to quantization and can be effectively combined without significant degradation.
\begin{table}[htbp]
\centering
\resizebox{0.48\textwidth}{!}{
\begin{tabular}{c|ccccc}
\toprule
\textbf{Method} & \textbf{PIQA} & \textbf{Winog} & \textbf{HellaSwag} & \textbf{ARC-E} & \textbf{ARC-C} \\
\midrule
Dense & 81.88 & 75.3 & 64.91 & 80.51 & 53.41 \\
FLAP & 77.56 & 61.08 & 55.23 & 70.36 & 42.37 \\
Ours & \textbf{78.45} & \textbf{72.85} & \textbf{57} & \textbf{75.8} & \textbf{46.67} \\
Ours + Quant & \textbf{78.21} & \textbf{71.98} & \textbf{56.88} & \textbf{75.21} & \textbf{46.1} \\
\bottomrule
\end{tabular}
}
\caption{Performance of our pruning method with 8-bit AWQ quantization on Qwen2.5-32B.}
\label{quant}
\end{table}

\section{Conclusion}
In this paper, we propose a pruning method that is deeply integrated with PD disaggregation. In particular, we construct pruning and distillation sets to perform iterative block removal, independently tailored to the prefill and decode, achieving better solutions compared to prior block pruning approaches. Moreover, we analyze the pruning sensitivity of the prefill and decode stages and identify removable blocks specific to each stage, making it well suited for PD disaggregation deployment. Under the same settings, our method achieves improved performance and faster inference. Extensive experiments demonstrate our approach consistently achieves strong performance in both PD disaggregation and PD unified (non-PD disaggregation) settings, and can also be extended to other non-block pruning methods.

\section*{Limitations}
In this work, we conduct extensive experiments to evaluate the effectiveness of our pruning method. The results demonstrate that our approach achieves competitive performance compared to the baselines. However, due to computational constraints, we have not yet been able to evaluate it on larger scale models, such as those with 70 billion parameters. Exploring the scalability of our method to such large models constitutes an important direction for future work.


\bibliography{custom}

\appendix
\clearpage

\section{Detailed Derivation of the Theoretical Formula}
\label{math_setp}

Denote by $X \in \mathbb{R}^{N \times d}$ the input sequence in the prefill stage and by $x_t \in \mathbb{R}^{1 \times d}$ the initial input in the decode stage. We denote the perturbations of the attention parameters ${W}_Q$, ${W}_K$ and ${W}_V$ by $\Delta {W}_Q$, $\Delta {W}_K$ and $\Delta {W}_V$, respectively. The query vector \(q\) and its perturbation \(\Delta q\) for $x_t$ are as follows:
\begin{equation}
q = x_t W_Q, \; 
\Delta q = x_t \Delta W_Q
\end{equation}
The concatenated key and value matrices \((K,V)\), along with their perturbations \((\Delta K,\Delta V)\), can be represented as follows:
\begin{equation}
K = [K_{pre},\, x_t W_K], \; 
V = [V_{pre},\, x_t W_V]
\end{equation}
\begin{equation}
\scalebox{0.81}{$
\Delta K = [\Delta K_{pre},\, x_t \Delta W_K], \; 
\Delta V = [\Delta V_{pre},\, x_t \Delta W_V]
$}
\end{equation}
Here, \( K_{{pre}} = X W_K \) and \( \Delta K = X \Delta W_K \); the same applies to the others. Then, the attention output before \((o)\) and after perturbation \((\tilde{o})\), as well as the resulting error \((E)\), can be expressed as follows:
\begin{equation}
o = {softmax}\left( \tfrac{q K^\top}{\sqrt{d}} \right) V
\end{equation}
\begin{equation}
\scalebox{0.95}{$
    \tilde{o} = {softmax}\\
\left( \tfrac{(q + \Delta q) (K + \Delta K)^\top}{\sqrt{d}} \right) (V + \Delta V)
$}
\end{equation}
\begin{equation}
\tilde{A} = {softmax}\left(\tfrac{(q + \Delta q)(K + \Delta K)^\top}{\sqrt{d}}\right)
\end{equation}
\begin{equation}
    \Delta A = \tilde{A} - {softmax}\left(\tfrac{q K^\top}{\sqrt{d}}\right)
\end{equation}
\begin{equation}
E = \tilde{o} - o 
= {\Delta A \, V} + {\tilde{A} \, \Delta V}
\end{equation}
To bound the Frobenius norm of the attention output perturbation $E$, we apply the triangle inequality to separate the contributions from the two terms, yielding the following expression:
\begin{equation}
\|E\|_F \le \|\Delta A \, V\|_F + \|\tilde{A} \, \Delta V\|_F
\end{equation}
Then, using the submultiplicativity of matrix norms, we can directly obtain the following bounds:
\begin{equation}
\|\Delta A \, V\|_F \le \|\Delta A\|_F \, \|V\|_F
\end{equation}
\begin{equation}
\|\tilde{A} \, \Delta V\|_F \le \|\tilde{A}\|_2 \, \|\Delta V\|_F
\end{equation}
By the Lipschitz continuity of the softmax, the perturbation $\Delta A$ can naturally be bounded as follows:
\begin{equation}
\scalebox{0.93}{$
\|\Delta A\|_F \le L_{{softmax}} \, \frac{\| (q + \Delta q)(K + \Delta K)^\top - q K^\top \|_F}{\sqrt{d}}
$}
\end{equation}
Here, $L_{{softmax}}$ is the Lipschitz constant of the softmax. Upon expanding the product, we can explicitly obtain the resulting expression as follows:
\begin{equation}
\begin{aligned}
(q + \Delta q)(K + \Delta K)^\top - q K^\top \\
= \Delta q \, K^\top  + q \, (\Delta K)^\top + (\Delta q)(\Delta K)^\top
\end{aligned}
\end{equation}
where the last term \((\Delta q)(\Delta K)^\top\) is of second order and can be neglected in a first order analysis. Applying the submultiplicativity of the Frobenius norm, we can further obtain the following bounds:
\begin{equation}
\|\Delta q \, K^\top\|_F \le \|\Delta q\|_2 \, \|K\|_F
\end{equation}
\begin{equation}
\|q (\Delta K)^\top\|_F \le \|q\|_2 \, \|\Delta K\|_F
\end{equation}
Considering all above analysis, we can express the Frobenius norm of the error $E$ as follows:
\begin{equation}
\begin{aligned}
\|E\|_F
&\le \frac{L_{{softmax}}}{\sqrt{d}}
\Big(
\|\Delta q\|_2 \, \|K\|_F
+ \|q\|_2 \, \\
&\|\Delta K\|_F \Big) \, \|V\|_F + \|\tilde{A}\|_2 \, \|\Delta V\|_F 
\end{aligned}
\end{equation}
Considering that $\|\tilde{A}\|_2 \le 1$, we obtain the following expression:
\begin{equation}
\begin{aligned}
\|E\|_F
&\le \frac{L_{{softmax}}}{\sqrt{d}}
\Big(
\|\Delta q\|_2 \, \|K\|_F
+ \|q\|_2 \, \\
&\|\Delta K\|_F
\Big)\cdot \|V\|_F  + \|\Delta V\|_F 
\end{aligned}
\end{equation}
Building on the above formula, and to more clearly illustrate the effect of pruning, we can express the error $E$ at step $t$ using an abstract error function $G$ as follows:
\begin{equation}
E = {G}(x_t, Y_{{pre}}, \Delta)
\end{equation}
where $Y_{{pre}}$ denotes the set of all results (including both intermediate and final outputs) produced in the previous step that can be reused in the current step, and $\Delta$ represents the set of perturbations.

\section{Frobenius Norm of \(\Delta A\) via Softmax Lipschitz Continuity}
\label{softmax}
For any $a, b \in \mathbb{R}^n$, the softmax function is Lipschitz continuous with respect to the Frobenius norm, which can be expressed as follows:
\begin{equation}
\label{soft_1_equ}
\scalebox{0.79}{$
\|{softmax}(a) - {softmax}(b)\|_F \le L_{{softmax}} \|a - b\|_F
$}
\end{equation}
Here, $L_{{softmax}}$ is the Lipschitz constant of the softmax. Based on the derivations in the main text, we obtain the following expression:
\begin{equation}
\label{soft_2_equ}
\begin{aligned}
\Delta A = softmax\Big(\tfrac{(q + \Delta q)(K + \Delta K)^\top}{\sqrt{d}}\Big) \\
- softmax\Big(\tfrac{q K^\top}{\sqrt{d}}\Big)
\end{aligned}
\end{equation}
By jointly considering Equations \ref{soft_1_equ} and \ref{soft_2_equ}, we can derive the following expression:
\begin{equation}
\scalebox{0.93}{$
\|\Delta A\|_F \le L_{softmax} \frac{\| (q + \Delta q)(K + \Delta K)^\top - q K^\top \|_F}{\sqrt{d}}
$}
\end{equation}

\section{Experimental Setup}
\label{Experimental Settings}

\textbf{Baselines }We conduct extensive comparative evaluations against other pruning algorithms. LLM-Pruner \citep{ma2023llm} adopts structural pruning that selectively removes non-critical coupled structures based on gradient information. FLAP \citep{an2024fluctuation} is a structured pruning framework that reduces storage by leveraging fluctuation based metrics and adaptive model compression. Shortened LLaMA \citep{kim2024shortened} selectively removes less important blocks based on block level importance scores, thereby accelerating model inference without significantly impacting performance. ShortGPT \citep{men2024shortgpt} defines a BI metric to measure the importance of each layer within the model and directly removes those layers. SLEB \citep{song2024sleb} employs a logit based approach to identify unnecessary transformer layers and updates the importance scores after each layer removal. SliceGPT \citep{ashkboos2024slicegpt} is a post-training sparsification scheme which replaces each weight matrix with a smaller matrix. Through these comprehensive comparisons, we thoroughly assess the strengths of our approach.

\textbf{Implementation Details }Our experiments are conducted using the PyTorch framework \citep{paszke2019pytorch} and the Hugging Face Transformers library \citep{wolf2020transformers}. We use two nodes within the same local area network, with each node equipped with one NVIDIA H100 80GB GPU. We follow the respective compression strategies for prefill and decode as mentioned above. We set the parameter \( d_T \) to 0.95 by default. During the iterative block selection process, we initialize the annealing temperature \( T_0 \) at 15 to ensure sufficient exploration in the early stages. The temperature is then gradually reduced with a decay coefficient \( \alpha = 0.85 \) until it reaches the minimum temperature \( T_{\min} = 0.05 \). When distilling two blocks, we use the weights of the block with low cosine similarity as the initial weights. We distill the block using the PIQA and MMLU as the training set. The training is performed using the Adam optimizer with a learning rate of \(1 \times 10^{-5}\), a batch size of 64 for 10 epochs. When sparsifying the KV Cache, we set the pruning ratio \( p = 0.3 \), with an attention score threshold \( \gamma = 0.75 \) to preserve the most semantically important key value pairs. We set the performance improvement threshold mentioned in Section \ref{Different Strategies for Prefill and Decode} to 3\%. We conduct experiments with various hyperparameters to demonstrate the stability of our method. By default, we use 256 randomly sampled examples from PIQA and MMLU as the calibration set. In the zero shot performance comparisons, we maintain consistent experimental settings, including the calibration datasets.

\section{Experiments on Strong Scalability and Robust Generality (Larger Model, More Datasets and Additional Metric)}
\label{Larger Model, More Datasets and Additional Metric}
We further evaluate our method on the larger Qwen2.5-32B model with a pruning ratio of 25\%. As reported in Table~\ref{32B}, our approach consistently achieves strong performance. Additionally, we report perplexity on WikiText2 using LLaMA2-13B and OPT-6.7B with pruning ratios of 24.37\% and 20\%, respectively, where our method achieves the best results (Table~\ref{perplexity}). Beyond these evaluations, we assess LLaMA3.1-8B-Instruct \citep{grattafiori2024llama} under a pruning ratio of 15.3\% on two further benchmarks: the instruction-following dataset IFEval \citep{zhou2023instruction} and the code generation dataset HumanEval \citep{chen2021evaluating}. As shown in Table~\ref{IFEval}, our method maintains strong performance across these tasks.
\begin{table}[htbp]
\centering
\resizebox{0.48\textwidth}{!}{
\begin{tabular}{c|ccccc}
\toprule
\textbf{Method} & \textbf{PIQA} & \textbf{Winog} & \textbf{HSwag} & \textbf{ARC-E} & \textbf{ARC-C} \\
\midrule
Dense & 81.88 & 75.3 & 64.91 & 80.51 & 53.41 \\
LLM-Pruner & 78.02 & 61.64 & 55.02 & 70.24 & 42.24 \\
FLAP & 77.56 & 61.08 & 55.23 & 70.36 & 42.37 \\
Shortened & 76.55 & 61.81 & 55.39 & 71.31 & 41.7 \\
ShortGPT & 76.66 & 72.53 & 52.4 & 72.85 & 42.24 \\
SLEB & 72.28 & 65.88 & 56.32 & 69.22 & 38.61 \\
 Ours & \textbf{78.45} & \textbf{72.85} & \textbf{57} & \textbf{75.8} & \textbf{46.67} \\
\bottomrule
\end{tabular}
}
\caption{Performance comparison of various methods on Qwen2.5-32B.}
\label{32B}
\end{table}

\begin{table*}[htbp]
\centering
\begin{tabular}{c|ccccccc}
\toprule
\textbf{Model} & \textbf{Dense} & \textbf{LLM-Pruner} & \textbf{FLAP} & \textbf{Shortened} & \textbf{ShortGPT} & \textbf{SLEB} & \textbf{Ours} \\
\midrule
OPT-6.7B & 10.86 & 11.92 & 11.68 & 12.51 & 13.68 & 12.94 & \textbf{11.28} \\
LLaMA2-13B & 4.88 & 8.16 & 9.73 & 10.81 & 9.25 & 8.96 & \textbf{7.38} \\
\bottomrule
\end{tabular}
\caption{Evaluation of perplexity on WikiText2 for LLaMA2-13B and OPT-6.7B with pruning.}
\label{perplexity}
\end{table*}

\begin{table*}[htbp]
\centering
\begin{tabular}{c|ccccccc}
\toprule
\textbf{Benchmark} & \textbf{Dense} & \textbf{LLM-Pruner} & \textbf{FLAP} & \textbf{Shortened} & \textbf{ShortGPT} & \textbf{SLEB} & \textbf{Ours} \\
\midrule
IFEval & 76.36 & 65.41 & 65.45 & 62.60 & 65.42 & 56.48 & \textbf{73.48} \\
HumanEval & 68.95 & 59.36 & 60.26 & 57.27 & 60.44 & 51.39 & \textbf{66.99} \\
\bottomrule
\end{tabular}
\caption{Evaluation of LLaMA3.1-8B-Instruct under 15.3\% pruning on IFEval and HumanEval.}
\label{IFEval}
\end{table*}

\section{Ablation Study}
\label{Ablation study}
\begin{figure}[htbp]                 
    \centering
    \includegraphics[width=1\linewidth]{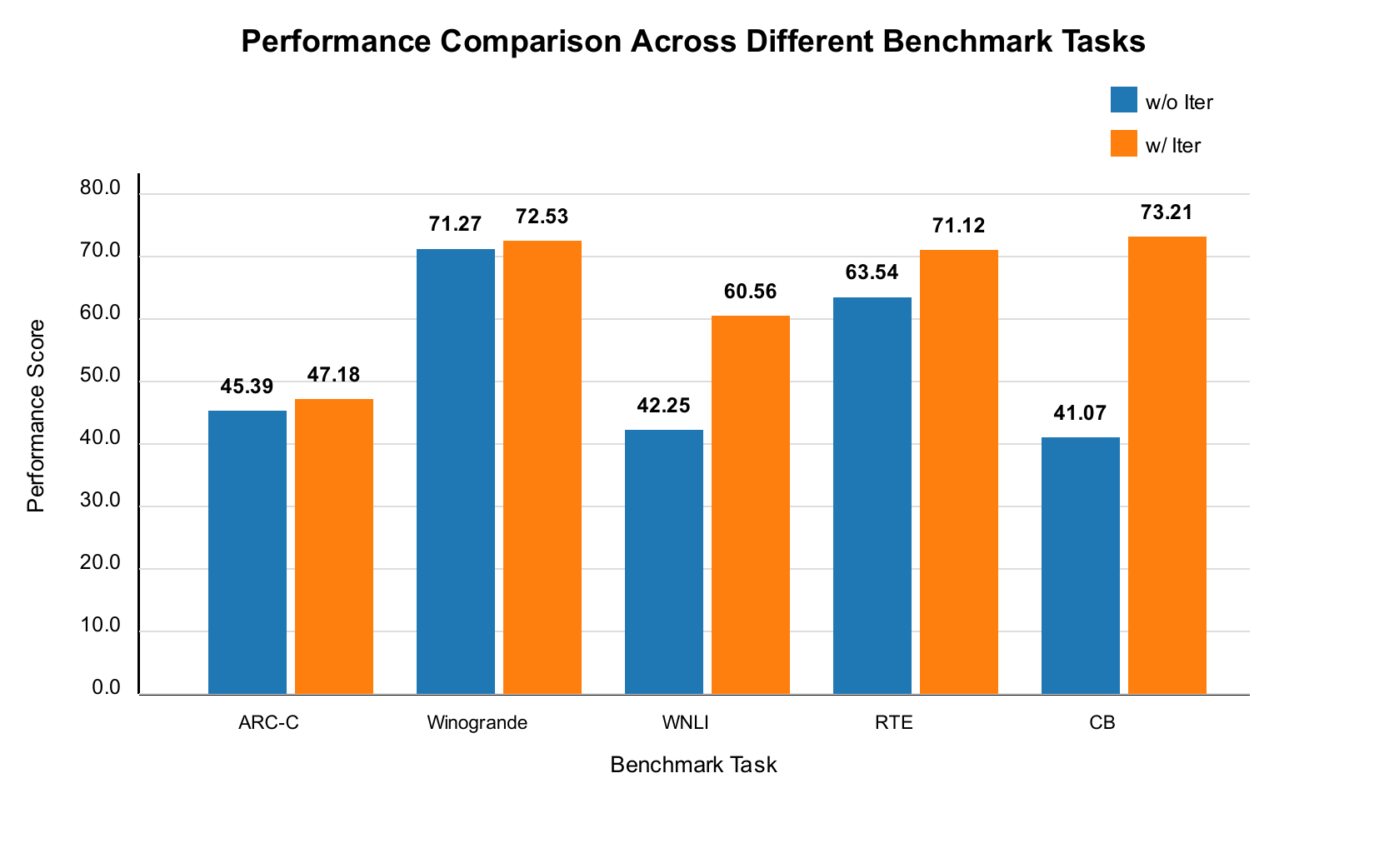}
    \caption{Ablation study on the effectiveness of the iterative block removal strategy: performance comparison with and without the strategy.}
    \label{fig:block_abl}
\end{figure}

We conduct comprehensive ablation studies to evaluate the individual contributions of the key component of our framework: the iterative block removal strategy. We conduct experiments on LLaMA3.1-8B, involving the pruning of 9.38\% blocks. The other hyperparameter settings used in these experiments are in the implementation details above. Block pruning performance, as shown in Figure \ref{fig:block_abl}, is significantly enhanced by the iterative optimization mechanism, which consistently outperforms its non iterative counterpart. This advantage stems from the iterative strategy's ability to more effectively navigate the search space, ultimately identifying superior block combinations that yield notable improvements in performance.

\section{Results of Different Pruning Ratios}
\label{Pruning Ratios}
\begin{figure}[htbp]
    \centering
    \includegraphics[width=1\linewidth]{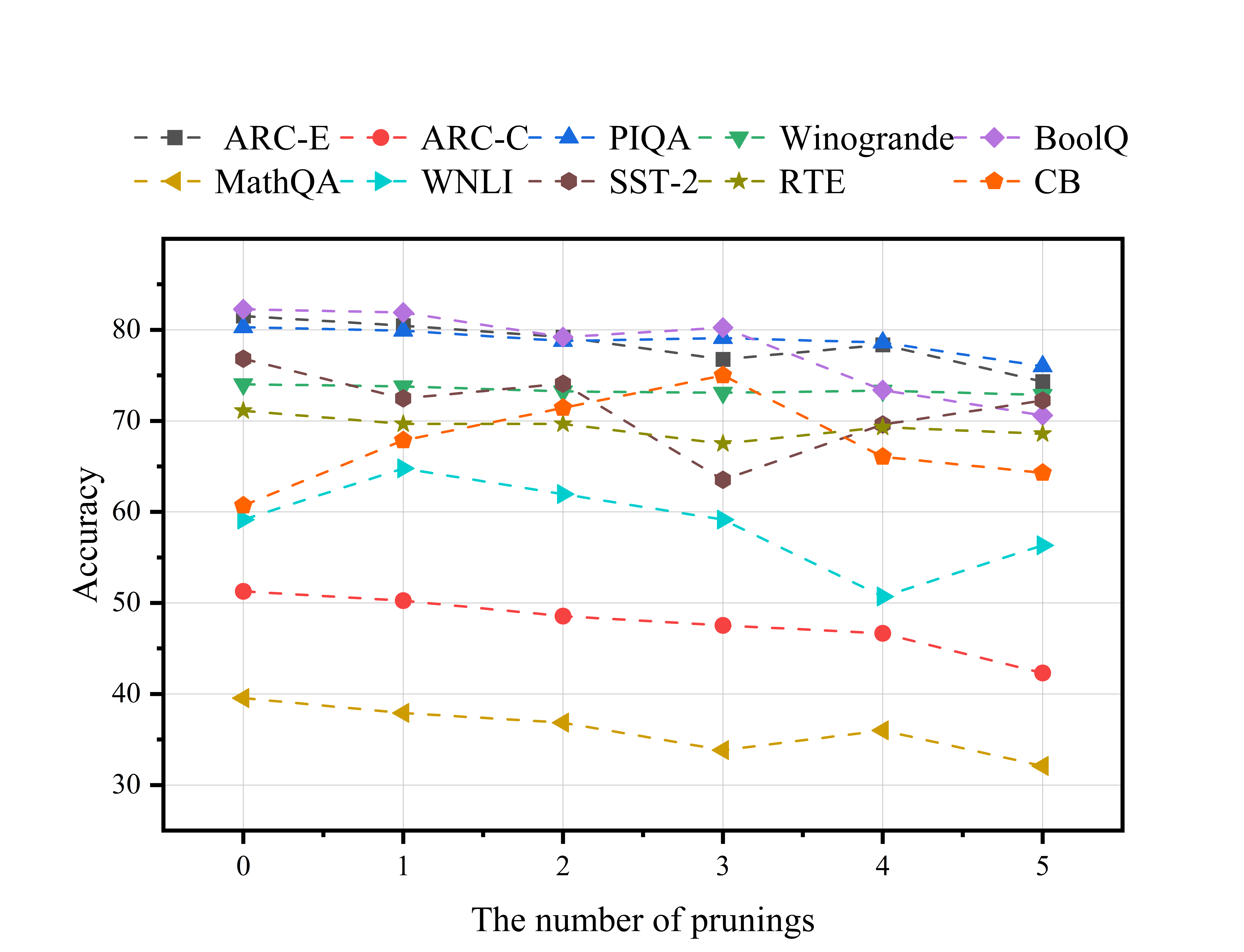}
    \caption{Results of model performance under different pruning blocks on LLaMA3.1-8B.}
    \label{fig:block_ratio}
\end{figure}

Figure \ref{fig:block_ratio} illustrates the impact of different pruning ratios on LLaMA3.1-8B performance. 
As the pruning ratio increases, the accuracy of most tasks remains relatively stable despite a general downward trend, indicating the strong robustness of our method. Block pruning accuracy remains stable even when the number of pruned blocks increases, especially on tasks such as PIQA, Winogrande and RTE. This phenomenon validates the effectiveness of our redundancy aware iterative block pruning strategy, which selectively removes non-critical blocks while preserving the model's core functional capabilities.

\begin{table*}[htbp] 
\setlength{\tabcolsep}{3.5pt} 
\centering 
\resizebox{\textwidth}{!}{%
\begin{tabular}{ccc|ccccccccccccc|cc} 
\toprule 
\textbf{$T$} & \textbf{$\alpha$} & \textbf{$T_{min}$} & \textbf{ARC-E} & \textbf{ARC-C} & \textbf{PIQA} & \textbf{Winog} & \textbf{BoolQ} & \textbf{MathQA} & \textbf{WNLI} & \textbf{SST-2} & \textbf{RTE} & \textbf{CB} & \textbf{MMLU} & \textbf{CMMLU} & \textbf{HSwag} & \textbf{AVG} \\ 
\midrule 
5 & 0.70 & 0.10 & 76.77 & 47.53 & 79.11 & 73.09 & 80.24 & 33.84 & 59.15 & 63.53 & 67.51 & 75.00 & 62.79 & 49.97 & 56.78 & 63.49 \\ 
10 & 0.80 & 0.10 & 78.11 & 47.53 & 78.62 & 71.82 & 79.08 & 36.31 & 61.97 & 57.45 & 72.56 & 73.21 & 63.43 & 49.91 & 56.33 & 63.56 \\ 
15 & 0.85 & 0.05 & 78.11 & 47.53 & 78.62 & 71.82 & 79.08 & 36.31 & 61.97 & 57.45 & 72.56 & 73.21 & 63.43 & 49.91 & 56.33 & 63.56 \\ 
\bottomrule 
\end{tabular}%
} 
\caption{Performance comparison across three key hyperparameter settings on the LLaMA3.1-8B. Under all these different parameter configurations, the model maintains a high level of performance.} 
\label{tab:block_para} 
\end{table*}

\begin{table*}[htbp]
\centering
\setlength{\tabcolsep}{3.5pt} 
\resizebox{\textwidth}{!}{
\begin{tabular}{c|ccccccccccccc|cc}
\toprule
\textbf{$d_T$} & \textbf{ARC-E} & \textbf{ARC-C} & \textbf{PIQA} & \textbf{Winog} & \textbf{BoolQ} & \textbf{MathQA} & \textbf{WNLI} & \textbf{SST-2} & \textbf{RTE} & \textbf{CB} & \textbf{MMLU} & \textbf{CMMLU} & \textbf{HSwag} & \textbf{AVG} \\ 
\midrule
0.97 & 77.44 & 47.18 & 78.40 & 72.53 & 76.82 & 36.82 & 60.56 & 59.75 & 71.12 & 73.21 & 58.74 & 48.39 & 55.38 & 62.80 \\
0.96 & 78.11 & 47.53 & 78.62 & 71.82 & 79.08 & 36.31 & 61.97 & 57.45 & 72.56 & 73.21 & 63.43 & 49.91 & 56.33 & 63.56 \\
0.95 & 77.65 & 47.18 & 78.40 & 73.40 & 79.57 & 36.21 & 64.79 & 69.72 & 73.29 & 76.79 & 60.76 & 43.37 & 57.22 & 64.49 \\
0.94 & 76.77 & 47.53 & 79.11 & 73.09 & 80.24 & 33.84 & 59.15 & 63.53 & 67.51 & 75.00 & 62.79 & 49.97 & 56.78 & 63.49 \\ 
\bottomrule
\end{tabular}
}
\caption{Evaluation of the influence of the block distillation threshold \( d_T \) on the LLaMA3.1-8B.}
\label{dT}
\end{table*}

\section{Hyperparameter Impact Analysis}
\label{hyperparameter}
We evaluate the impact of key hyperparameters in our iterative block removal strategy, including the initial temperature \( T \), temperature decay coefficient \( \alpha \), and minimum temperature \( T_{{min}} \). We prune 9.38\% blocks on LLaMA3.1-8B. As shown in Table \ref{tab:block_para}, increasing \( T \) and \( \alpha \) from the first configuration to the second configuration improves the average accuracy from 63.49 to 63.56. This indicates that a slower cooling schedule helps more effectively explore the solution space and avoid converging to suboptimal local minima. Importantly, further increasing the parameter values in the third configuration does not change the performance across all benchmarks, suggesting that the pruning process has converged. This convergence implies that the algorithm has identified a near optimal block removal combination, beyond which additional iterations or a larger search space will not yield further gains. Overall, all three configurations are able to find a high quality solution space. Additionally, we further assess the threshold parameter \( d_T \) for constructing the distillation set. We prune 9.38\% blocks of the LLaMA3.1-8B model. As shown in Table~\ref{dT}, the experimental results indicate that different threshold parameters can all achieve relatively optimal outcomes, consistently demonstrating good performance.

\section{Analysis of Iterative Block Removal Solutions}
\label{Analysis of Iterative Block Removal Solutions}
To validate the effectiveness of our iterative block removal method, we compare its solutions with the global optimum obtained via exhaustive search. It is important to note that exhaustive search requires evaluating all possible block combinations, which is practically intractable. To make this comparison feasible, we select all elements from the distillation set and the top five elements from the pruning set, and treat the optimal solution among all their combinations as the global optimum. Even so, solving it still demands substantial computational time. We apply our method to prune 13.6\% of the parameters in LLaMA3.1-8B and evaluate it across various benchmarks. As shown in Table~\ref{tab:optium}, our approach achieves results consistent with the global optimum. This demonstrates that our iterative method efficiently finds superior pruning solutions.
\begin{table}[h]
\centering
\resizebox{0.48\textwidth}{!}{
\begin{tabular}{c|ccccc}
\toprule
\textbf{Method} & \textbf{PIQA} & \textbf{Winog} & \textbf{HSwag} & \textbf{ARC-E} & \textbf{ARC-C} \\
\midrule
Global Optimum & 76.99 & 73.56 & 53.80 & 73.15 & 43.86 \\
Ours           & 76.99 & 73.56 & 53.80 & 73.15 & 43.86 \\
\bottomrule
\end{tabular}}
\caption{Performance of iterative block removal compared to the global optimum on LLaMA3.1-8B.}
\label{tab:optium}
\end{table}

\section{Analysis of Using Cosine Similarity}
\label{Analysis of Using Cosine Similarity}

Besides cosine similarity, dot product and Euclidean distance are also commonly used, but they are sensitive to vector magnitude. Previous work \citep{chen2024streamlining} has shown that in Transformers, hidden states tend to expand with depth, leading deeper layers to exhibit higher dot-product similarity, while shallower layers maintain smaller Euclidean distances. To avoid this bias, we use cosine similarity, which is insensitive to magnitude. In addition, we measure the L2 norms of the hidden states at each layer of Qwen2.5-7B. As shown in Table \ref{tab:hs_norms}, the norm magnitude exhibits a clear upward trend with increasing depth, which highlights the inherent bias of other similarity metrics that are sensitive to vector magnitude. Furthermore, we evaluate the performance of Qwen2.5-7B and OPT-13B under 15\% parameter pruning with different similarity metrics. As shown in Table \ref{tab:metric_perf}, cosine similarity consistently outperforms the other metrics.
\begin{table}[htbp] 
    \centering
    \resizebox{0.48\textwidth}{!}{
    \begin{tabular}{c|ccc}
        \toprule
        \textbf{Metric} & \textbf{PIQA} & \textbf{Winog} & \textbf{HSwag} \\
        \midrule
        Euclidean distance & 75.70 & 52.20 & 50.51 \\
        Dot product        & 75.87 & 55.88 & 47.62 \\
        Cosine similarity  & 77.04 & 56.99 & 51.19 \\
        \bottomrule
    \end{tabular}}
        \caption{Performance comparison on Qwen2.5-7B using different similarity metrics.}
    \label{tab:metric_perf}
\end{table}

\begin{table*}[htbp]
    \centering
    \begin{tabular}{c|cccccccccccccc}
        \toprule
        \textbf{Layer} & 0 & 1 & 2 & 3 & 4 & 5 &$\cdots$ & 21 & 22 & 23 & 24 & 25 & 26 & 27 \\
        \midrule
        \textbf{Norm}  & 14 & 19 & 25 & 116 & 141 & 146 & $\cdots$ & 264 & 300 & 350 & 408 & 470 & 474 & 620 \\
        \bottomrule
    \end{tabular}
        \caption{The norms of the hidden states output at each layer of Qwen2.5-7B. 
    The first row represents the layer IDs, and the second row represents the corresponding norms.}
    \label{tab:hs_norms}
\end{table*}

\begin{table*}[htbp]
\centering
\begin{tabular}{c|c|cc}
\toprule
\textbf{Calibration Set} & \textbf{Size} & \textbf{15.63\% Blocks Removal} & \textbf{9.38\% Blocks Removal} \\ 
\midrule
\multirow{3}{*}{PIQA} & 256 & \multirow{3}{*}{24, 25, 26, [22, 23], [27, 28]} & \multirow{3}{*}{25, 26, [27, 28]} \\
 & 512 &  &  \\
 & 1024 &  &  \\ 
\midrule
\multirow{3}{*}{WikiText-2} & 256 & \multirow{3}{*}{24, 25, 26, [22, 23], [27, 28]} & \multirow{3}{*}{25, 26, [27, 28]} \\
 & 512 &  &  \\
 & 1024 &  &  \\ 
\bottomrule
\end{tabular}
\caption{Results comparison across different calibration sets. Under the LLaMA3.1-8B model, we report the selected block indices for removal across different calibration datasets and various sizes under two pruning configurations: removing 15.63\% blocks and removing 9.38\% blocks. The indices within square brackets denote the distilled blocks.}
\label{tab:stable}
\end{table*}

\section{Robustness on Different Calibration Sets}
\label{Calibration Sets}
To evaluate the stability of our proposed block pruning strategy, we systematically analyze the consistency of results across different calibration sets and sizes. Our analysis distinguishes between two types of removal operations: direct block pruning and block distillation. Direct block pruning refers to the removal of blocks, while block distillation involves merging blocks instead of removing them. As shown in Table \ref{tab:stable}, the experimental results demonstrate significant stability in the selection of removed blocks across all experimental conditions. This consistency strongly indicates that our pruning decisions are driven by the inherent architectural redundancy within the model itself, exhibiting robustness across different datasets.

\section{More Details on Prefill and Decode Removal Schemes}
\label{More_Details_on_Prefill_and_Decode_Removal_Schemes}
Our approach first determines the optimal removal strategy and then derives separate removal schemes for the prefill and decode stages. Specifically, the prefill scheme is designed as a subset of the decode scheme, ensuring that every retained block in the decode node can directly reuse the corresponding KV Cache from the prefill node. In contrast, if separate removal schemes are determined for the prefill and decode stages at the initial stage, without any subset relationship, it is likely that some blocks retained in the decode node would have no corresponding KV Cache in the prefill node, hindering cache reuse and reducing system efficiency.

\section{Comparison with Work of Similar Names}
\label{Comparison with Work of Similar Names}
We compare our approach with another method \citep{muralidharan2024compact} that bears a similar name, as both involve concepts such as iteration and distillation. In their work, iteration refers to repeatedly performing pruning to gradually reduce the model size, whereas in our approach, iteration is used to search for the optimal pruning configuration, and the two notions are not directly related. Similarly, their distillation is applied after pruning to the entire model in order to restore performance, while our distillation occurs during the pruning process itself, serving as a mechanism for block removal.

\section{Case Study}
\label{Case Study}
To qualitatively assess the impact of pruning on generation quality, we compared the outputs of the original model and the pruned model for several representative prompts, as shown in Table \ref{tbl:case}.  We prune 9.38\% blocks on LLaMA3.1-8B. Despite a substantial reduction in model size, the pruned model consistently generates coherent, information-rich, and well-structured responses, retaining the core semantics of the original output. These examples confirm that our pruning strategy maintains strong generation capabilities and factual consistency even under significant structural compression.

\begin{table*}[htbp]
    \centering
    \resizebox{\linewidth}{!}{
    \begin{tabular}{p{0.5\textwidth}|p{0.5\textwidth}}
        \toprule
        \multicolumn{1}{c|}{Original Model} & \multicolumn{1}{c}{Pruned Model} \\
        \midrule
        \underline{Explain what artificial intelligence is.} Artificial intelligence (AI) is the simulation of human intelligence processes by machines, especially computer systems. These processes include learning (the acquisition of information and rules for using the information), reasoning (using the rules to reach approximate or definite conclusions), and self-correction. Particular applications of AI ... & \underline{Explain what artificial intelligence is.} Artificial intelligence, or AI, is a field of computer science that deals with the creation of machines that can perform tasks that normally require human intelligence. This includes things such as understanding natural language, recognizing objects in images, and making decisions based on complex data. While AI has been around for decades, recent ... \\
        \midrule
        \underline{Provide some tips for learning a new language.} 1. Set realistic goals: Start by setting realistic goals for yourself, such as learning a few basic phrases or mastering a specific skill. 2. Practice regularly: Consistency is key when it comes to learning a new language. 3. Find a language partner: Finding a language partner can be a great way to practice and improve your language skills. 4. Use technology... & \underline{Provide some tips for learning a new language.} Here are some tips for learning a new language: 1. Start by learning the basics of the language, such as the alphabet, common words and phrases, and basic grammar rules. 2. Find a language learning app or program that suits your learning style and goals. 3. Practice speaking and listening to the language as much as possible. 4. Practice ... \\
        \midrule
        \underline{Explain how a car engine works.} A car engine is a machine that converts the chemical energy stored in gasoline into mechanical energy that can be used to move the car. The engine consists of several parts that work together to produce this energy. The first part of the engine is the combustion chamber. This is where the gasoline is mixed with air and ignited to create an explosion. The explosion pushes down on a piston, which is connected to a crankshaft. The crankshaft is what turns wheels ... & \underline{Explain how a car engine works.} A car engine works by converting chemical energy stored in fuel into mechanical energy that drives the car. The process of converting chemical energy into mechanical energy is known as combustion. In a car engine, combustion occurs in the combustion chamber, which is located in the center of the engine. The combustion chamber is connected to the intake manifold, which is where the fuel and air mixture enters the engine. The intake manifold ... \\
        \bottomrule
    \end{tabular}
    }
    \caption{Qualitative examples comparing outputs from the original and the pruned versions of the LLaMA3.1-8B models configured as described above. Each example illustrates the model’s response to the same input prompt, with the underlined text indicating the user’s question.} \label{tbl:case}
\end{table*}

\end{document}